\documentclass[11pt,a4paper]{article}
\usepackage[hyperref]{naaclhlt2019}
\usepackage{times}
\usepackage{latexsym}
\usepackage{graphicx}
\usepackage{amsmath}
\usepackage{algorithm}
\usepackage[noend]{algpseudocode}

\usepackage{url}

\aclfinalcopy 

\title{Medical Multimodal Classifiers Under Low Data Situations}

 \author{Faik Aydin$^{\clubsuit}$ \hspace{5mm} 
\hspace{3mm}
 Maggie Zhang$^{\spadesuit}$ 
\hspace{3mm} 
 Michelle Ananda-Rajah$^{\heartsuit,\clubsuit}$ 
\hspace{3mm} 
 Gholamreza Haffari$^{\clubsuit,\diamondsuit}$ \\ \\
  $\clubsuit$: Monash University \\
 $\spadesuit$: NVIDIA \\
 $\heartsuit$: Alfred Hospital \\
 $\diamondsuit$: {\tt gholamreza.haffari@monash.edu}
 }

\begin{document}
\maketitle
\begin{abstract}

Data is one of the essential ingredients to power deep learning research. Small datasets, especially specific to medical institutes, bring challenges to deep learning training stage. This work aims to develop a practical deep multimodal that can classify patients into abnormal and normal categories accurately as well as assist radiologists to detect visual and textual anomalies by locating areas of interest. The detection of the anomalies is achieved through a novel technique which extends the integrated gradients methodology with an unsupervised clustering algorithm. This technique also introduces a tuning parameter which trades off true positive signals to denoise false positive signals in the detection process. To overcome the challenges of the small training dataset which only has 3K frontal X-ray images and medical reports in pairs, we have adopted transfer learning for the multimodal which concatenates the layers of image and text submodels. The image submodel was trained on the vast ChestX-ray14 dataset, while the text submodel transferred a pertained word embedding layer from a hospital-specific corpus.  Experimental results show that our multimodal improves the accuracy of the classification by 4\% and 7\% on average of 50 epochs, compared to the individual text and image model, respectively.   
\end{abstract}

\section{Introduction}

The field of medical imaging, was augmented with the introduction of the CNN (Convolutional Neural Network). State of the art results demonstrate deep learning in medical imaging area could reach the ability of radiologist level \citep{chx:17}. These could be achieved if there are massive datasets to train the deep CNNs. However, the challenge in practice is that most medical datasets are small, domain specific, and restricted to medical institutes. Therefore, the motivation of this work is to handle small data situation for classification and detection of anomalies in medical imaging and reports.

We present a multimodal, which jointly takes medical reports and the corresponding images as input, to extract all the relevant information in small data environment. Apart from an image submodel, a 1-D CNN based text submodel inspired by the research done for e-commerce space \citep{esk:17} is developed for anomaly classification. Transfer learning is also applied to take advantage of large open source datasets, in order to improve the accuracy of both image and text parts. 

In addition to classification, our model can be used for detection in both the images and the reports, providing easily intepretable highlights of the anomalies to assist medical experts and patients. Our detection method also addresses the problem of structured noise present in the image detection process. Structured noise can be defined as false positive signals carried due to transfer learning. False positive signals are eliminated via trading off some true positive signals which can yield a clean detection.

\section{The Multimodal}

There are few existing research using multimodal for medical datasets. State of the art works \cite{mag:1,mag:2,mag:3,mag:4} focus on using multimodal to generate standard medical reports, while our work can produce classification and detection results on both medical reports and images.

This section focuses on our novel multimodal architecture which consists of a text submodel and an image submodel, as shown in Figure 1. The text submodel is developed by taking the network layers, from input layers to feature vectors, from a trained text classifier. Similarly, the image submodel takes the network layers of a trained image classifier from input layers to feature vectors. Then the encoded text and image feature vectors are concatenated into a single flat feature vector. This feature vector is then passed onto a simple densely connected decoder for the binary
classification. Applying transfer learning to the two encoders  makes this multimodal function under small data situations. The transferred text and image encoders are the respective pre-trained embedding and residual layers. The pre-training process is described in sections 2.1 and 2.2. These pretrained encoders are then finely tuned on low learning rates. 

\subsection{Text Submodel}

We choose a CNN based text classifier \citep{Yoon-Kim-2014} to meet this task. We also extend the 1-D CNN for classifying short e-commerce product descriptions \citep{esk:17} to our text classifier. Resemblance in structure of product descriptions, features of short radiology reports (27.4 words per case on average) are extracted by the Word2Vec approach \citep{mik:13}. Each document is cut off in 140 words, or padded if a given document is shorter than 140 words.
In the spirit of transfer learning, different pre-trained word embedding layers are experimented on, and these are reported in the experimental results (see Table 1 and 2). The eventual text classifier architecture utilizes pre-trained embedding layers specific to the text dataset that is being classified. The choice of a domain specific embedding layer rather then a generalized one is to achieve a clear boost in performance. As stated earlier, the corpus is domain specific and institution specific. Kernels used in the 1-D CNN are of length 3,4 and 5 and the pre-trained embedding layer is not frozen.

\begin{figure}
\includegraphics [width = 3in]{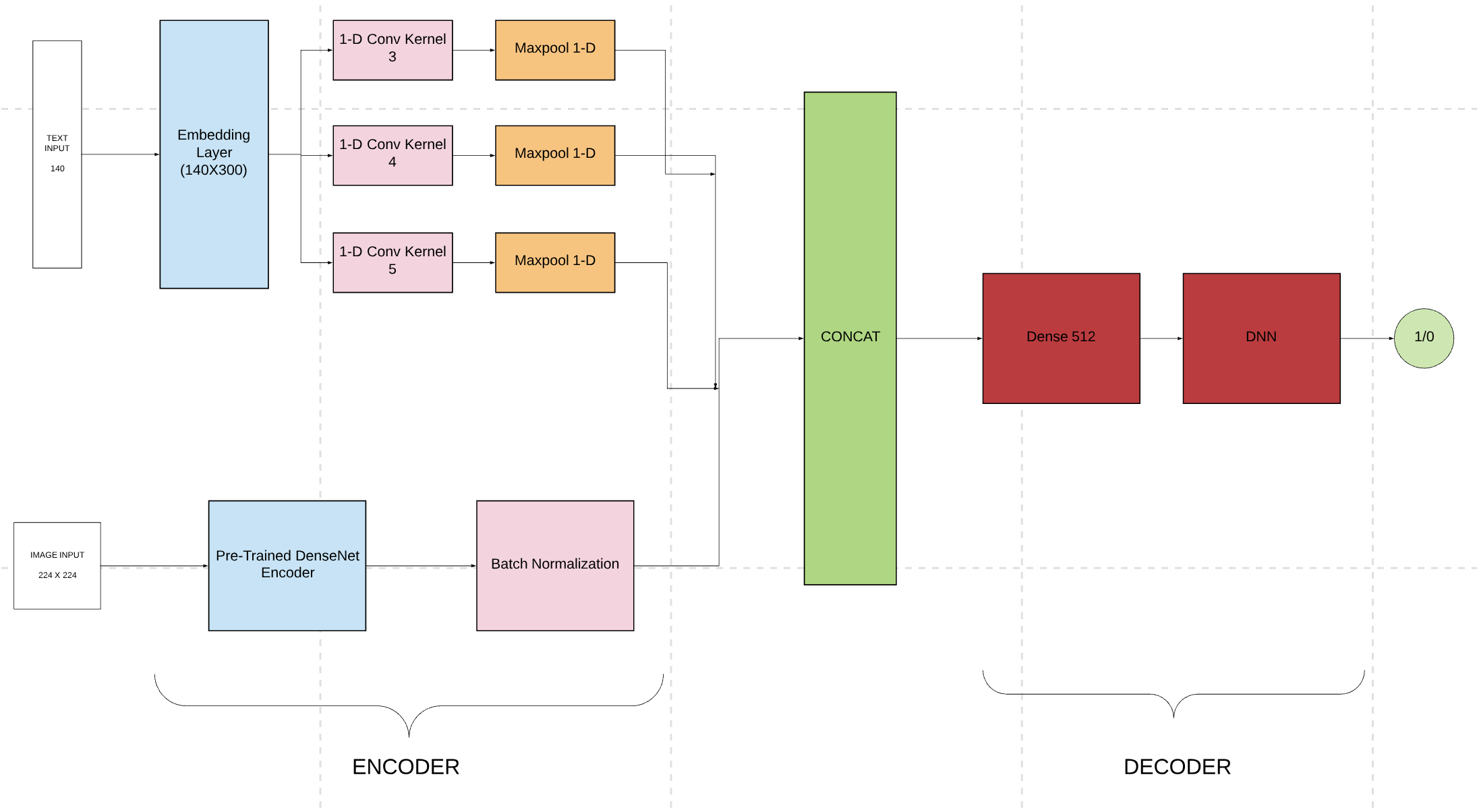}
\label{fig:my_label}
\caption{Multimodal Architecture. Encoder top half: text submodel. Encoder bottom half: image submodel}
\end{figure}

\subsection{Image Submodel}

The image submodal is centered around the idea of transfer learning. CNN encoders pre-trained on ImageNet and National Health Institute’s ChestX-ray14 gave several different experimentation combinations with diferent encoders (VGG, DenseNet, ResNet, etc.). The DenseNet121 \citep{he:15} pre-trained on ChestX-ray14 was chosen as the encoder due to its superior accuracy. This encoder is then fed into a simple batch normalization which is then fed into a simple image decoder for binary classification.

\section{Detection}

\subsection{Image Detection: Unsupervised Integrated Gradients}

Detection in transfer learning has its challenges of structured noise. This noise can be defined as false positive signals. The source of this problem is that the encoder is trained on a large dataset. The knowledge from this is transferred to a target for detection. Along with disease patterns, the encoder also carries features it learned from the larger dataset that are not areas of interest. These could be a dark background due to a lack of standard in torso placement or standard writing on the X-rays. The transfer cannot differentiate between the areas of interest and the noise and gives off positive signals for both.

We address this issue by introducing a tuning parameter called \textit{sight sensitivity}. Treating the image as a 2-D grid, this algorithm treats positive signals as points on a plane. These signals are obtained via the integrated gradients. These partial gradients of the image yield the influence of each pixel on the resulting classification decision. A large partial gradient value of a pixel is regarded as a positive signal. The \textit{sight sensitivity} parameter is the threshold value which decides if a given gradient is large enough to be considered as a signal or not.

This parameter can be tuned to yield signal points on the 2-D grid. All other pixel gradients that fail to surpass this threshold are turned off (0 values). The remaining points are clustered in an unsupervised manner. Based on information loss criteria of distance, the points are clustered around their anomaly neighborhoods. The centroids of these clusters act as the center of the circles drawn around these bounding circles. The radius of the bound is equal to the farthest away point's distance to its cluster's center point.

Above methodology is depicted in Figure 2. The ground truth for the patient is: "There are degenerative changes in the spine. Borderline enlarged heart.". The detection in the low sight sensitivity setting (bottom right) lets too much signal pass through. The result of this is detailed bounds around the spinal area and with an emphasis on the heart, accompanied with noise on the bottom left portion of the image. The higher sight sensitivity (bottom left) produces an averaged out explanation without the structured noise by trading off some true positive signals.

\begin{figure}[h]
\centering
\includegraphics [width = 3in]{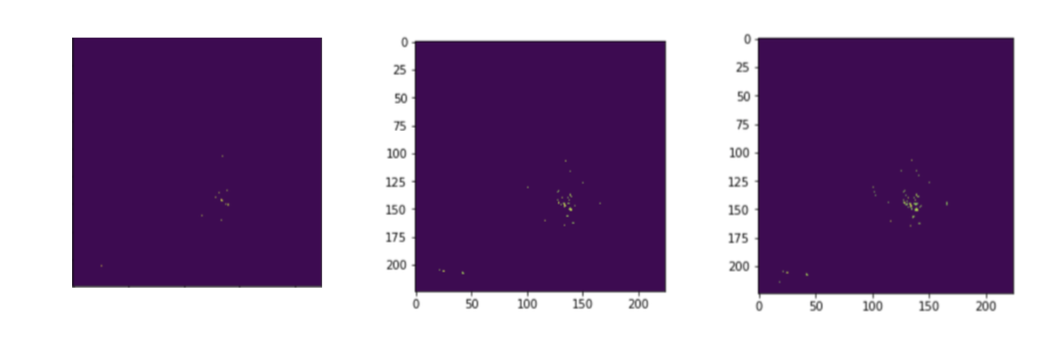}
\includegraphics [width = 3in]{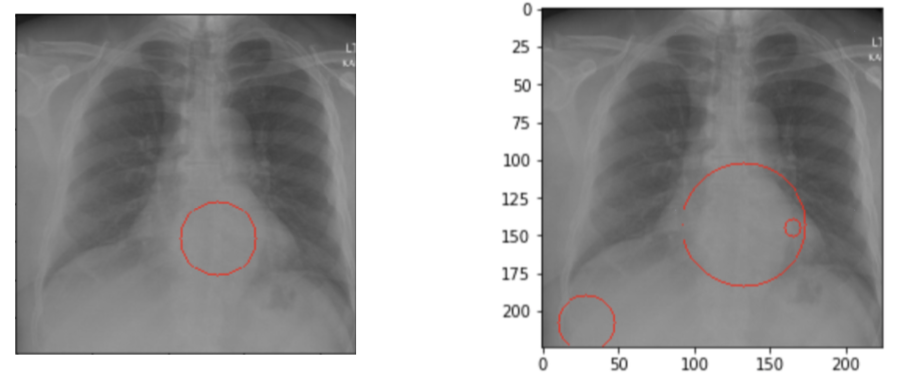}
\label{fig:my_label}
\caption{Noise Reduction. Top left to right: How signals are portrayed as points as sight sensitivity goes from high (left) to moderate (middle) to low (right). Bottom Left: High Sight Sensitivity, Bottom Right: Low Sight Sensitivity}
\end{figure}

\subsection{Text Explanations}
 Taking the gradients of inputs yields decent explanations for image networks. We utilize this tool for text explanations. Given that each word is treated as a 300 length array, a sentence can be thought of as a N by 300 matrix (or picture), where each row is a word. The gradient of each cell in the matrix are square summed per row. This yields a cumulative gradient score for each word. The normalization of this score vector basically gives the percentage of effect a word had in the overall score. Color coding these words with respect to their importance score makes it easier to zoom into the relevant - brighter colored - areas of the text. Figure 3 has a block of text converted into this readable explanation format. The top figure is the entirety of the text. Zooming into the lighter shaded area (bottom figure), shows us the note "there are degenerative changes of the spine", with the word degenerative shows a high indicator for abnormality (score of 0.72, with an arrow pointing to the word "degenerative"). Keep in mind that these are not scores per word, but per word instance in the sequence. This means the word degenerative can have a score of 0.4 in one sentence but may have 0.8 in another within the same document.

\begin{figure}[h]
\centering
\includegraphics [width = 3in]{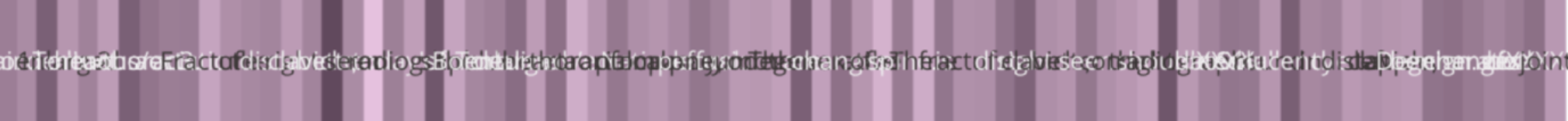}
\includegraphics [width = 3in]{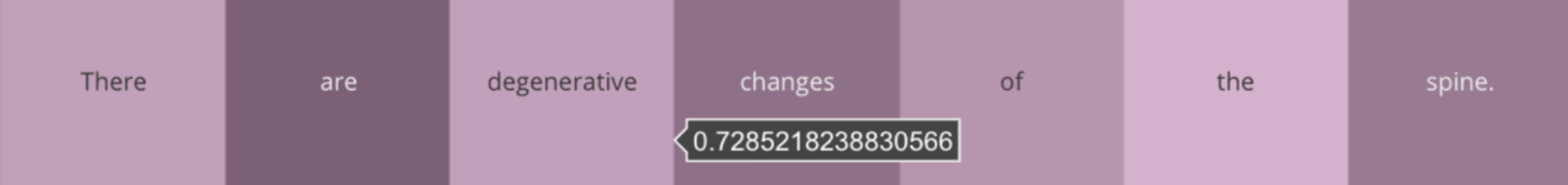}
\label{fig:my_label}
\caption{text explanations, top: entire document, bottom: important area of the document }
\end{figure}

\section{Experiments}
 Small datasets become a serious problem in the medical space when it comes to text. Unlike image, there are few quality open data sources. Therefore it becomes important to test any text classifier's performance on both an open data source and also a practical industry source. For this reason, there are two target dataset used for the text submodel. The open data source used for this work comes from Indiana University \citep{dem:15} (which we will refer to as the Indiana University dataset from this point on). The Indiana University X-ray image dataset which holds valuable text meta-data in the form of radiologist  notes (3955 cases, 60 \% of which are abnormal). The private dataset comes from Alfred Hospital (Melbourne, Victoria), again in the form of radiologist  notes (3000 cases, 60 \% of which are abnormal).

Different pre-trained embedding layers were used to test the domain specificity of this classification. The GloVe embedding layer was used as a generic embedding while custom embedding layers were developed for Alfred and The Indiana University dataset. The results for the experiments can be seen in table 2.  

\begin{table}[h]
\begin{tabular}{llll} 
  Embedding & Training & Validation & Testing\\
  \hline
  Custom & 0.88 & 0.80 & 0.76 \\
  GloVe & 0.73 & 0.73 & 0.68
\end{tabular}
\caption{Text model Accuracy on Alfred Hospital Dataset}
\end{table}

\begin{table}[h]
\begin{tabular}{llll}
  Embedding & Training & Validation & Testing\\
  \hline
  Custom & 0.88 & 0.80 & 0.75 \\
  GloVe & 0.72 & 0.71 & 0.69
\end{tabular}
\caption{Text model Accuracy on Indiana University Dataset}
\end{table}

Transfer learning is the crux of the image model. As we have done previously with the embedding layer in the text model, this work checks for generic feature performance in the image model as well. The generic feature's performance is assessed by transferring encoders trained on ImageNet and domain specific features are assessed with encoders trained on Chestxray-14. Densenet121 \citep{Huang:17} is chosen to be used as the encoder due to dense connections. Figure 4 shows the performance of these two different transfer learned models on the Indiana University dataset's validation set during fine tuning.

\begin{figure}[h]
\centering
\includegraphics [width = 3in]{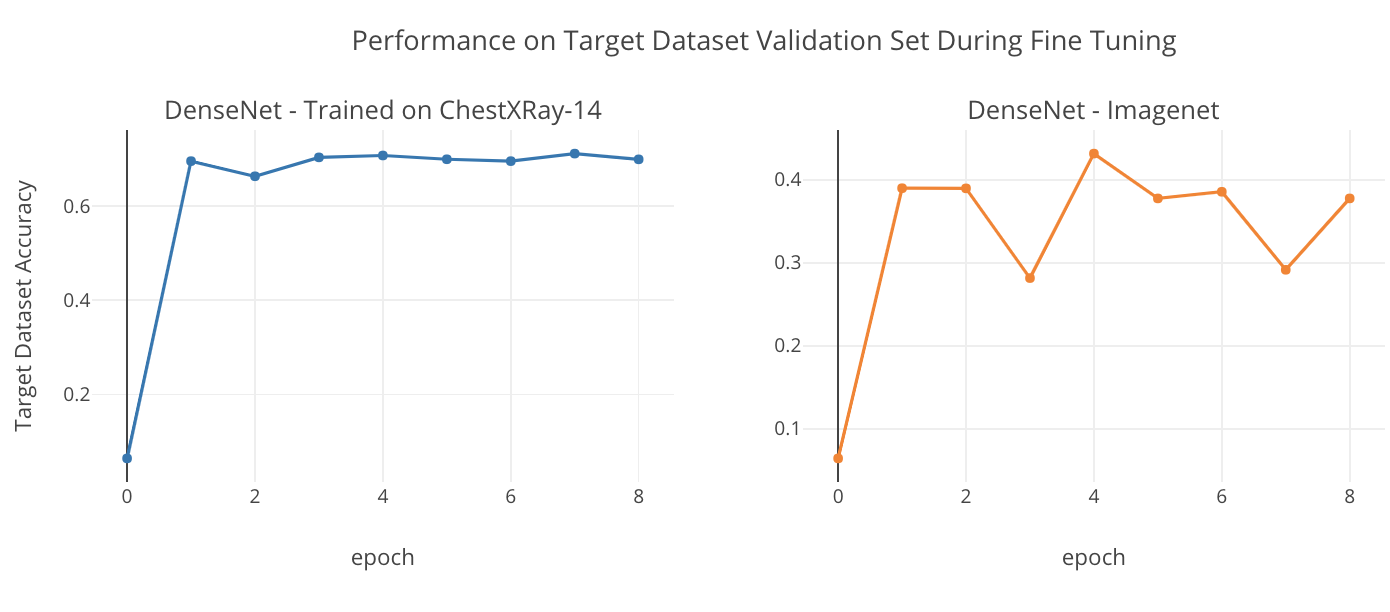}
\label{fig:my_label}
\caption{Transfer learning performance on target validation set}
\end{figure}

The target dataset for multimodal classification is the Indiana University \citep{dem:15} X-ray image dataset. Only frontal X-ray images and their respective radiologist  notes were used for this work.

The multimodal on average improves accuracy efficiently by 4\% and 7\% compared to the baseline models which are an individual text model and an image model when fine tuned with learning rate $10^{-5}$. As show in Figure 5, shows the performance of our multimodal compared to two baseline models which are individual text model and image model on Indiana University dataset tuned with a learning rate of $10^{-2}$. The lines represent the mean accuracy and the spread represents the variance over the course of 10 stratified splits and 10 epochs. The bottom graph shows accuracy after tuning with a learning rate of $10^{-5}$. This yields a total dominance of the multimodal over the course of 50 epochs.  

The experiments of over 10 stratified splits of the data yielded the following average results for multimodal and its stand alone submodals with high learning rates over 10 epochs. The low learning rate experiments were done with 3 stratified splits over 50 epochs.

\begin{table}[h]
\begin{tabular}{llll}
  Learning Rate & Multimodal & Text & Image \\
  \hline
  Low & 0.77 & 0.74 & 0.66 \\
  High & 0.81 & 0.77 & 0.74
\end{tabular}
\caption{Multimodal vs Baseline Models Accuracy}
\end{table}

\begin{figure}[h]
\centering
\includegraphics [width = 3in]{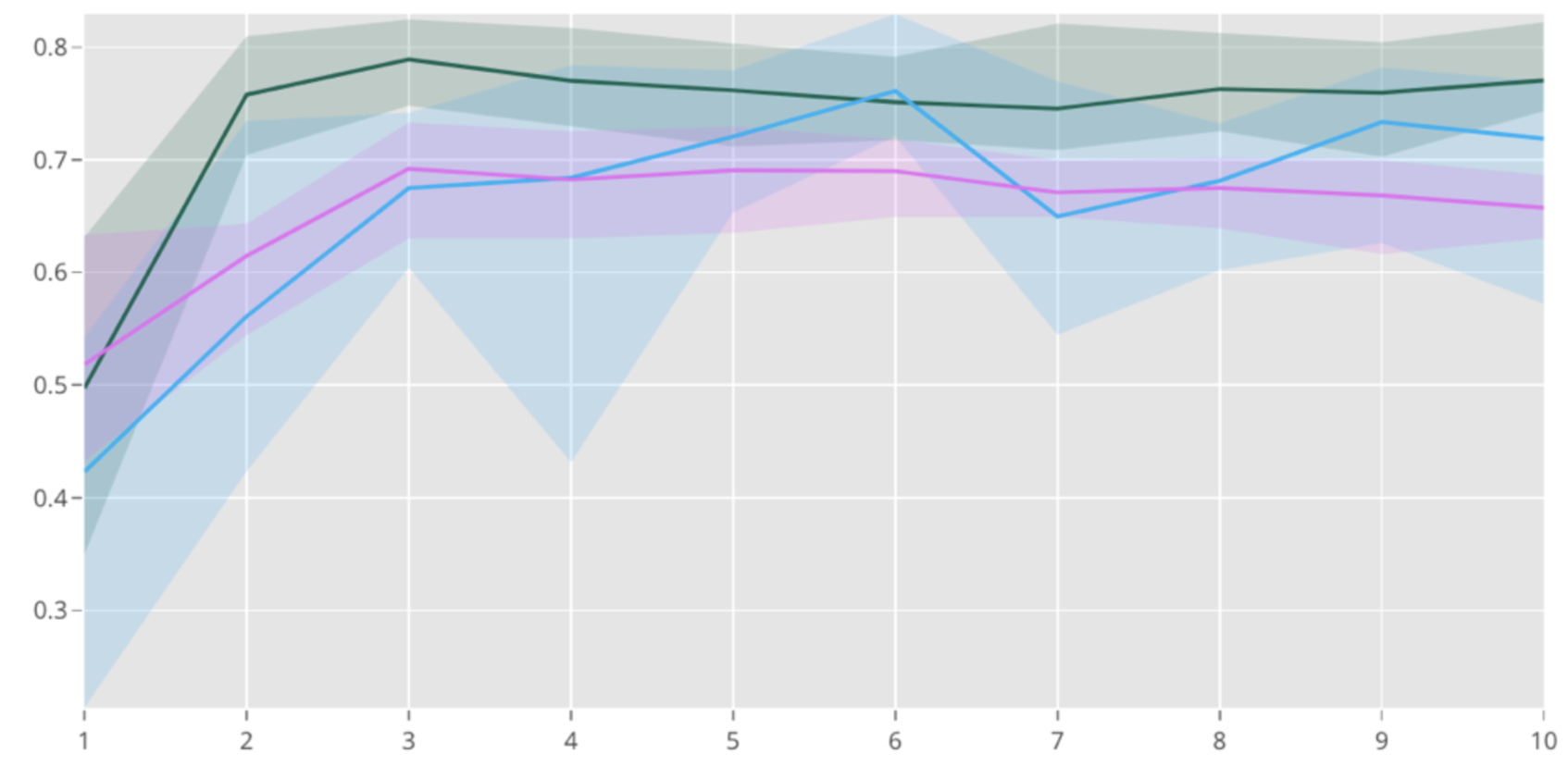}
\includegraphics [width = 3in]{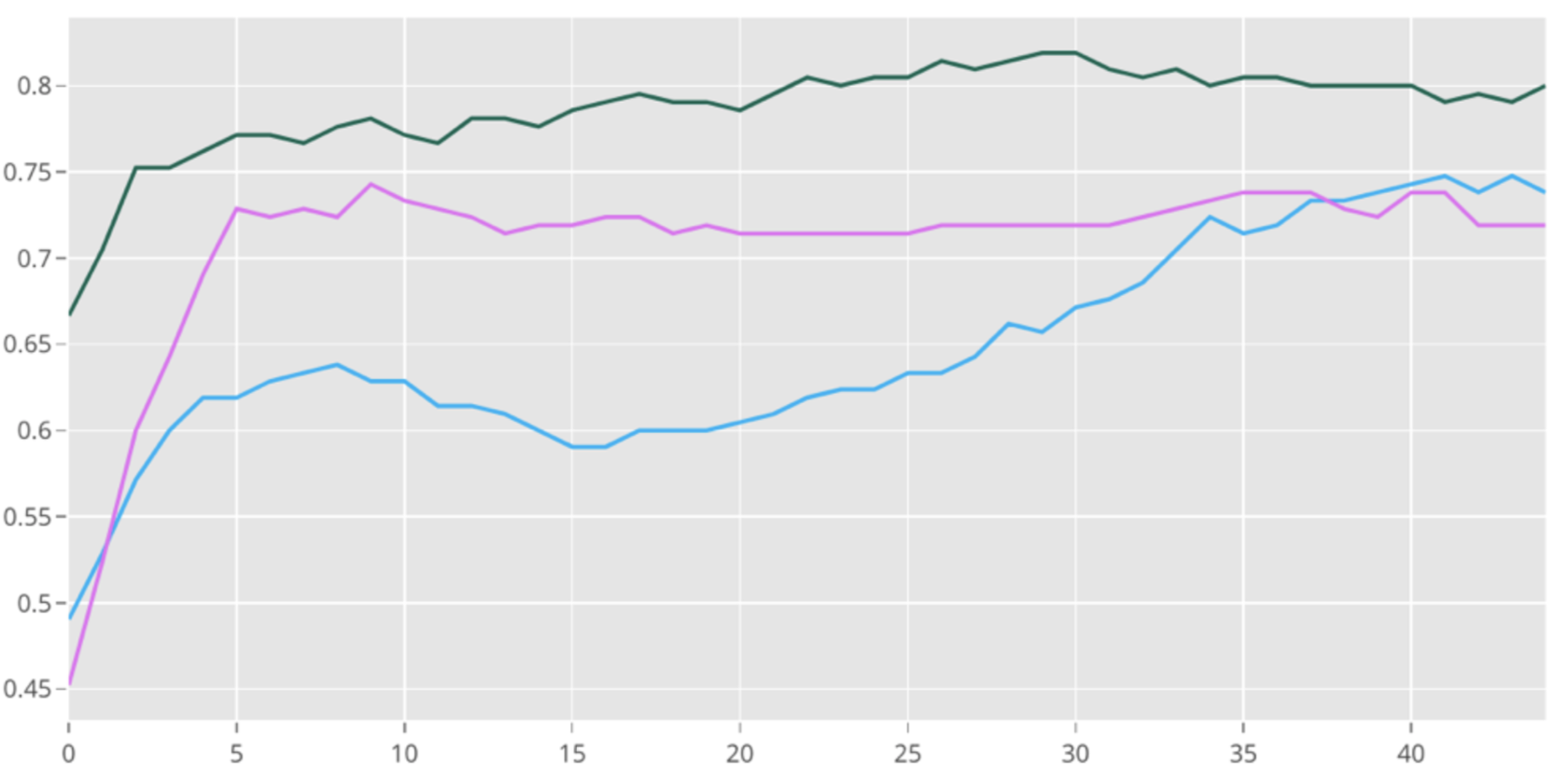}
\label{fig:my_label}
\caption{model comparison. top: high learning rate, bottom: low learning rate. accuracy vs epoch.
Green: Multimodal Classifier, Pink: Image Classifier, Blue: Text Classifier}
\end{figure}

\section{Conclusion}

This work aims to overcome the challenges of medical classification in the low data scenario. The use of multimodals and transfer learning proved to be promising and receive higher performance then stand alone image and text models. Treating sentences as images and taking pure gradients yielded proper explanations. Expanding the integrated gradients approach with unsupervised clustering gave noise free localized detection in the image context.  Future works will include different use cases of multimodals in the medical context, along with the use of attention models for images in order to avoid noise problems with transfer learning.

\bibliography{naaclhlt2019}

\begin{thebibliography}{11}
\expandafter\ifx\csname natexlab\endcsname\relax\def\natexlab#1{#1}\fi

\bibitem[{Demner-Fushman and Dina(2015)}]{dem:15}
Demner-Fushman and Marc B. Rosenman Sonya E. Shooshan Laritza Rodriguez Sameer
  Antani George R. Thoma Clement J.~McDonald Dina, Marc D.~Kohli. 2015.
\newblock Preparing a collection of radiology examinations for distribution and
  retrieval.
\newblock \emph{Journal of the American Medical Informatics Association},
  23(2):304--310.

\bibitem[{Eskesen(2017)}]{esk:17}
Sophie Eskesen. 2017.
\newblock Improving product categorization by combining image and title.

\bibitem[{He et~al.(2015)He, Z., R., and S.}]{he:15}
Kaiming He, Xiangyu Z., Shaoqing R., and Jian S. 2015.
\newblock \href {https://arxiv.org/pdf/1512.03385.pdf} {Deep residual learning
  for image recognition}.

\bibitem[{Huang(2017)}]{Huang:17}
Liu Z. Van Der Maaten L. Weinberger~K.Q Huang, G. 2017.
\newblock \href {https://arxiv.org/pdf/1608.06993.pdf} {Densely connected
  convolutional networks}.
\newblock arXiv:1608.6993.

\bibitem[{Kim(2015)}]{Yoon-Kim-2014}
Yoon Kim. 2015.
\newblock \href {https://arxiv.org/pdf/1408.5882.pdf} {Convolutional neural
  networks for sentence classification}.
\newblock arXiv:1408.5882.

\bibitem[{Mikolov(2013)}]{mik:13}
Sutskever I. Chen K. Corrado G. S. Dean~J Mikolov, T. 2013.
\newblock Distributed representations of words and phrases and their
  compositionality.
\newblock \emph{Advances in neural information processing systems}, pages
  3111--3119.

\bibitem[{Moradi et~al.(2017)Moradi, Mehdi, Madani, Gur, Guo, , and
  Syeda-Mahmood.}]{mag:4}
Moradi, Mehdi, Ali Madani, Yaniv Gur, Yufan Guo, , and Tanveer Syeda-Mahmood.
  2017.
\newblock Bimodal network architectures for automatic generation of image
  annotation from text.
\newblock \emph{International Conference on Medical Image Computing and
  Computer-Assisted Intervention}, pages 449--456.

\bibitem[{Rajpurkar et~al.(2017)Rajpurkar, Irvin, Zhu, Yang, Mehta, Duan, Ding,
  Bagul, Langlotz, Shpanskaya, Lungren, and Ng}]{chx:17}
Pranav Rajpurkar, Jeremy Irvin, Kaylie Zhu, Brandon Yang, Hershel Mehta, Tony
  Duan, Daisy Ding, Aarti Bagul, Curtis Langlotz, Katie Shpanskaya, Matthew~P.
  Lungren, and Andrew~Y. Ng. 2017.
\newblock \href {https://arxiv.org/abs/1711.05225} {Chexnet: Radiologist-level
  pneumonia detection on chest x-rays with deep learning}.
\newblock arXiv:1711.5225.

\bibitem[{Wang et~al.(2018)Wang, Xiaosong, Peng, Lu, Lu, and Summers}]{mag:1}
Wang, Xiaosong, Yifan Peng, Le~Lu, Zhiyong Lu, and Ronald~M. Summers. 2018.
\newblock Tienet: Text-image embedding network for common thorax disease
  classification and reporting in chest x-rays.
\newblock \emph{IEEE Conference on Computer Vision and Pattern Recognition},
  pages 9049--9058.

\bibitem[{Zhang et~al.(2017{\natexlab{a}})Zhang, Zizhao, Chen, Sapkota, and
  Yang}]{mag:2}
Zhang, Zizhao, Pingjun Chen, Manish Sapkota, and Lin Yang. 2017{\natexlab{a}}.
\newblock Tandemnet: Distilling knowledge from medical images using diagnostic
  reports as optional semantic references.
\newblock \emph{International Conference on Medical Image Computing and
  Computer-Assisted Intervention}, pages 302--328.

\bibitem[{Zhang et~al.(2017{\natexlab{b}})Zhang, Zizhao, Xie, Xing, McGough, ,
  and Yang}]{mag:3}
Zhang, Zizhao, Yuanpu Xie, Fuyong Xing, Mason McGough, , and Lin Yang.
  2017{\natexlab{b}}.
\newblock Mdnet: A semantically and visually interpretable medical image
  diagnosis network.
\newblock \emph{IEEE conference on computer vision and pattern recognition},
  pages 6428--6436.

\end{thebibliography}
\bibliographystyle{acl_natbib}

\end{document}